\documentclass[a4paper,fleqn]{cas-dc}

\usepackage[numbers,sort&compress]{natbib}
\usepackage{float}

\usepackage{balance}
\usepackage{threeparttable}

\def\tsc#1{\csdef{#1}{\textsc{\lowercase{#1}}\xspace}}
\tsc{WGM}
\tsc{QE}
\tsc{EP}
\tsc{PMS}
\tsc{BEC}
\tsc{DE}

\begin{document}
\let\ref\Cref 		
\let\eqref\Cref 	
\let\autoref\Cref 	
\let\WriteBookmarks\relax
\def\floatpagepagefraction{1}
\def\textpagefraction{.001}

\footmarks{}

\bookmark[named = FirstPage]{Multimodal Joint Prediction of Traffic Spatial-Temporal Data with Graph Sparse Attention Mechanism and Bidirectional Temporal Convolutional Network} 

\title [mode = title]{Multimodal Joint Prediction of Traffic Spatial-Temporal Data with Graph Sparse Attention Mechanism and Bidirectional Temporal Convolutional Network}    




\author[1]{Dongran Zhang}
\ead{zhangdr5@mail2.sysu.edu.cn}

\address{Guangdong Provincial Key Laboratory of Intelligent Transportation System,  School of Intelligent Systems Engineering, Sun Yat-sen University,  Shenzhen 518107, China} 

\author[1]{Jiangnan Yan}
\ead{yanjn3@mail2.sysu.edu.cn}

\author[2]{Kemal Polat}
\ead{kpolat@ibu.edu.tr}
\address{Department of Electrical and Electronics Engineering, Faculty of Engineering, Bolu Abant Izzet Baysal University, Bolu, Turkey} 

\author[3]{Adi Alhudhaif}
\ead{a.alhudhaif@psau.edu.sa}
\address{Department of Computer Science, College of Computer Engineering and Sciences, Prince Sattam bin Abdulaziz University, Al-Kharj, 11942, Saudi Arabia} 

\author[1]{Jun Li}
\cormark[1]
\cortext[cor1]{Corresponding author}
\ead{stslijun@mail.sysu.edu.cn}

\begin{abstract} 
Traffic flow prediction plays a crucial role in the management and operation of urban transportation systems. While extensive research has been conducted on predictions for individual transportation modes, there is relatively limited research on joint prediction across different transportation modes. Furthermore, existing multimodal traffic joint modeling methods often lack flexibility in spatial-temporal feature extraction. To address these issues, we propose a method called Graph Sparse Attention Mechanism with Bidirectional Temporal Convolutional Network (GSABT) for multimodal traffic spatial-temporal joint prediction. First, we use a multimodal graph multiplied by self-attention weights to capture spatial local features, and then employ the Top-U sparse attention mechanism to obtain spatial global features. Second, we utilize a bidirectional temporal convolutional network to enhance the temporal feature correlation between the output and input data, and extract inter-modal and intra-modal temporal features through the share-unique module. Finally, we have designed a multimodal joint prediction framework that can be flexibly extended to both spatial and temporal dimensions. Extensive experiments conducted on three real datasets indicate that the proposed model consistently achieves state-of-the-art predictive performance.

\end{abstract}



\received { 27 December 2023}
\revised { 8 April 2024}
\accepted { 29 March 2024}
\online { 18 April 2024}


\maketitle

\section{Introduction}
\label{SE:Inroduction}

Short-term traffic flow prediction is an important component of intelligent transportation systems and has great potential for improving transportation system efficiency and passenger travel experience \cite{ye2021coupled}. From the perspective of urban computing, modern cities encompass various modes of transportation, such as taxis, shared bicycles, and subways. These modes generate diverse transportation data, and the spatial-temporal data from different cities often exhibit similar patterns of change, especially during peak traffic hours \cite{yang2023short}. Multimodal joint prediction integrates information from multiple data sources, utilizes spatial-temporal features of different traffic modes, enhances prediction accuracy and generalization, and effectively supports optimization and decision-making in intelligent transportation systems \cite{xu2023multi}.

In the past, short-term traffic flow prediction primarily focused on specific transportation modes. Early studies mainly utilized statistical models such as Auto Regressive Integrated Moving Averages (ARIMA) \cite{van1996combining}, Kalman filtering  \cite{okutani1984dynamic}, and exponential smoothing \cite{williams1998urban}. These models are more suitable for smooth traffic flow data. With the development of machine learning methods, techniques such as Support Vector Machine (SVM) \cite{cortes1995support, feng2018adaptive}, k-Nearest Neighbor (KNN) \cite{cai2016spatiotemporal}, and Gradient Boosting Decision Trees (GBDT) \cite{ma2017prioritizing} have also been applied to non-linear traffic flow prediction. These methods can adapt to non-linear traffic flow data, but they rely on manually set features and find it challenging to make multi-step predictions.

In recent years, deep learning-based methods have been widely applied in spatial-temporal traffic prediction \cite{zheng2023hybrid, cheng2022long, ye2024mvts}. Convolutional Neural Networks (CNN) \cite{zhang2017deep}, Graph Convolutional Networks (GCN) \cite{guo2019attention, ye2022learning, zhuo2022efficient, zhuo2024partitioning}, and spatial attention mechanisms \cite{chen2022bidirectional} are commonly used for spatial feature extraction, while Long Short-Term Memory networks (LSTM) \cite{ye2019co, ye2023multi}, Gated Recurrent Unit (GRU) \cite{ye2021coupled, cheng2022long}, temporal attention mechanisms \cite{zheng2020gman, jiang2023fecam}, and Temporal Convolutional Networks (TCN) \cite{bai2018empirical, zhang2023spatial, zhang2023multi} are commonly used for temporal feature extraction. These methods are capable of extracting complex spatial-temporal features from various perspectives, which in turn enhances prediction performance and generalization ability. However, their main limitation lies in their focus on a single mode of traffic, often neglecting the potential mutual influence between different traffic modes \cite{liang2022joint}. From a broader perspective, there is a strong correlation between various modes of transportation, both spatially and temporally \cite{ye2019co}. For instance, buses and subways exhibit a certain degree of substitution effect \cite{ding2016predicting}, while subways and shared bicycles commonly display characteristics of connectivity and inter-modal transportation \cite{lv2021mobility}. Thus, implementing multimodal prediction can provide more effective guidance for operational strategies and enhance the flexibility of the traffic system.

Although some research has made progress in predicting specific traffic modes and in joint modeling of multimodal traffic, there is still much room for improvement.

\begin{figure*}
	\centering
	\caption{Characteristics of multimodal traffic spatial-temporal data}
	
	\begin{subfigure}{0.49\textwidth}
		\centering
		\includegraphics[width=0.7\linewidth]{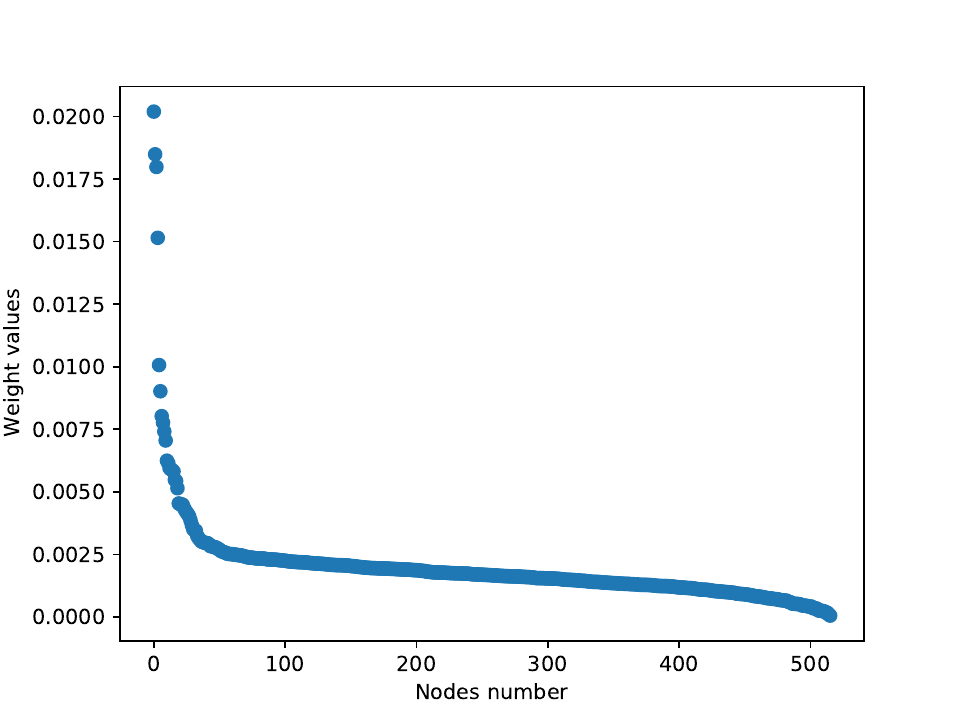}
		\caption{Long tailed distribution of self-attention weights.}
		\label{Fig: LongTail}
	\end{subfigure}
	\hfill
	\begin{subfigure}{0.49\textwidth}
		\centering
		\includegraphics[width=0.9\linewidth]{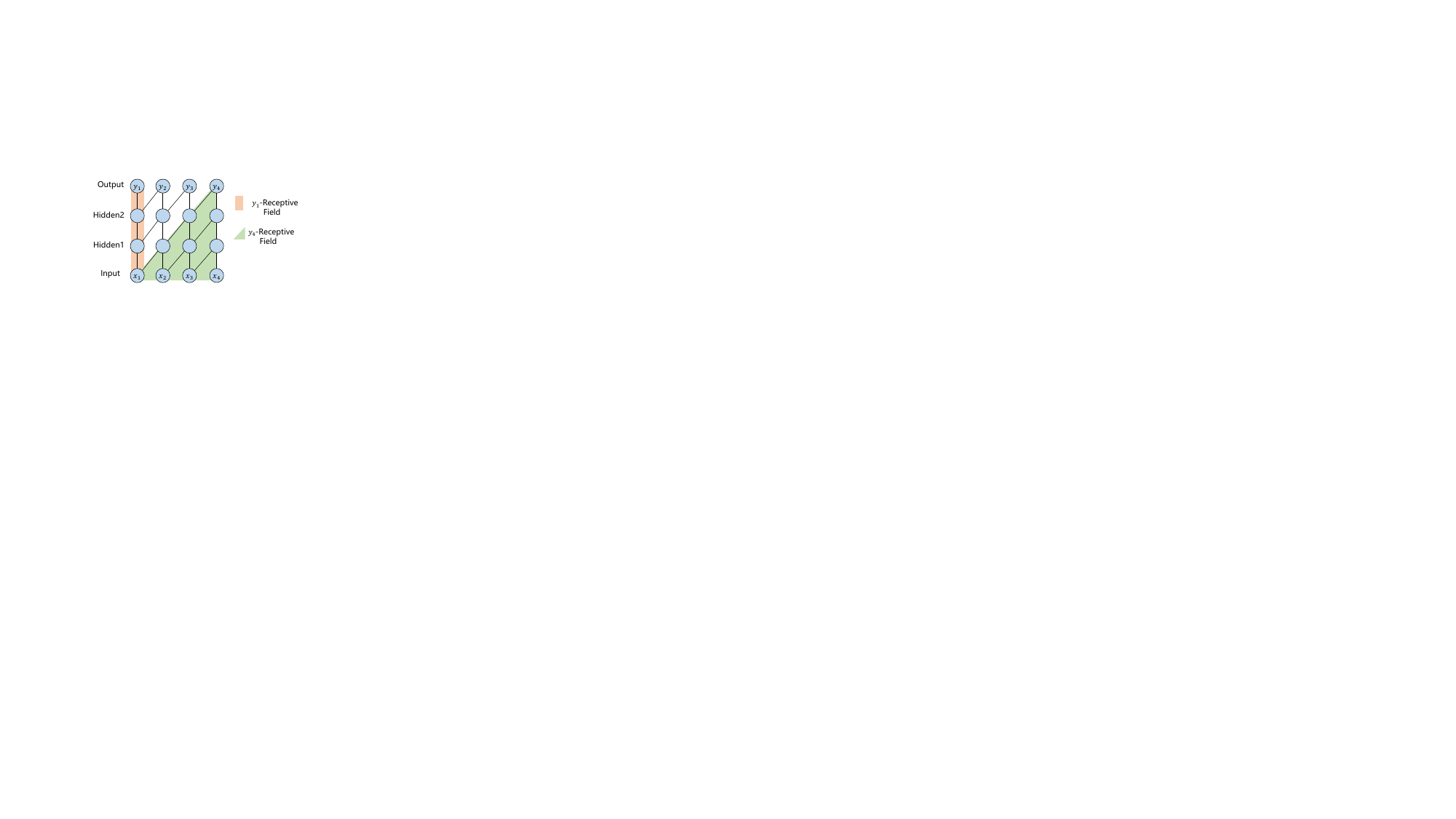}
		\caption{Limitations of TCN feature extraction.}
		\label{Fig: Tcn_Limit}
	\end{subfigure}
	\begin{subfigure}{0.49\textwidth}
		\centering
		\includegraphics[width=0.7\linewidth]{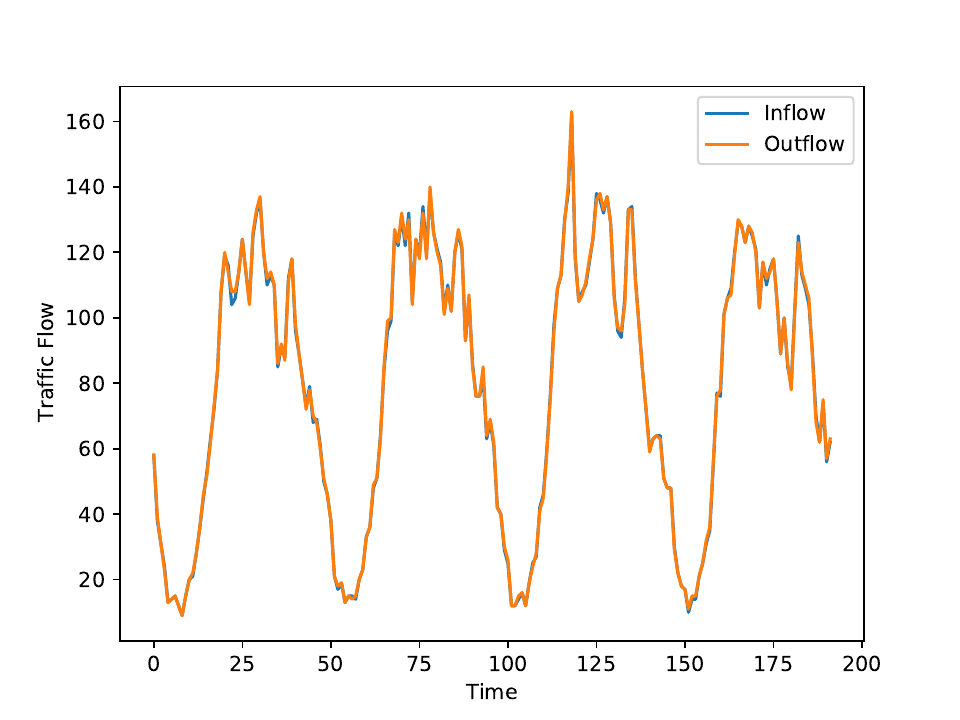}
		\caption{Inflow and Outflow traffic of a certain area in BJ Taxi.}
		\label{Fig: BJFeature}
	\end{subfigure}
	\hfill
	\begin{subfigure}{0.49\textwidth}
		\centering
		\includegraphics[width=0.7\linewidth]{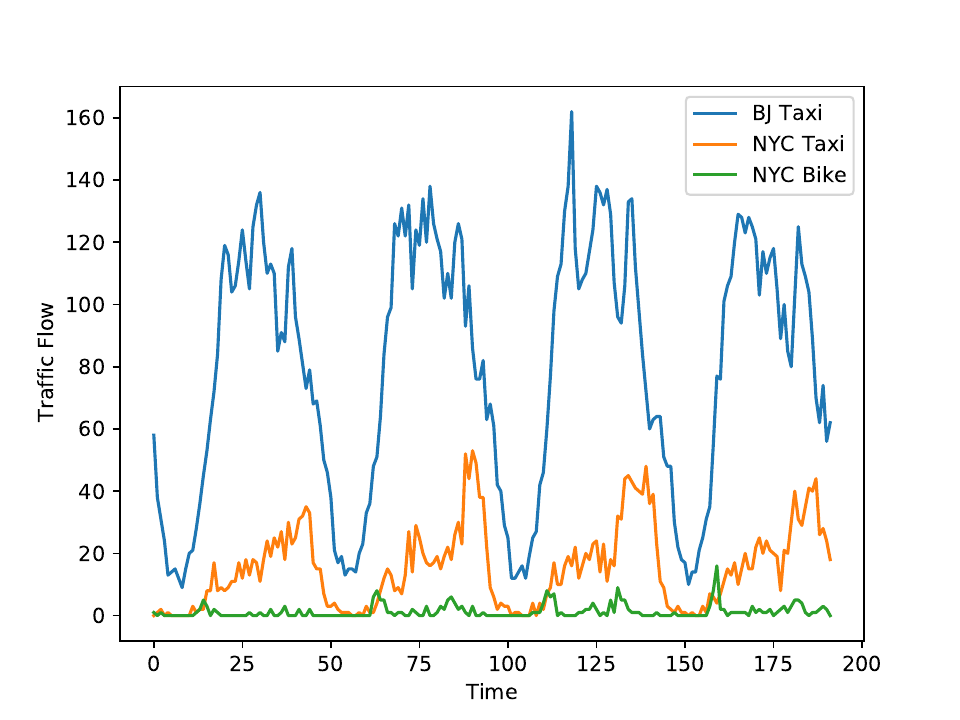}
		\caption{The scale differences of traffic flow in different modals.}
		\label{Fig: Multimode_Feature}
	\end{subfigure}

	\label{fig:2x2subfigures}
\end{figure*}

\begin{itemize}
	
	\item In terms of spatial feature extraction, the self-attention mechanism has emerged as a powerful tool capable of extracting global features. However, an inherent limitation of this mechanism is its indiscriminate inclusion of all nodes in the computational process. This approach deviates from actual traffic conditions and can potentially induce a 'long tail' effect, as depicted in Figure \ref{Fig: LongTail}. This effect can interfere with the extraction of weights from nodes of significant importance.

	\item In terms of temporal feature extraction, the Temporal Convolutional Network (TCN) has emerged as an effective model. However, it does have certain limitations, primarily due to the use of dilated causal convolution. This operation results in the output data of the TCN being unable to obtain the position information of the corresponding input data, thus failing to fully capture the temporal correlation between the input and output data. As shown in \ref{Fig: Tcn_Limit}, $ \{x_1, x_2, x_3, x_4\}$ constitute the inputs, while $ \{y_1, y_2, y_3, y_4\}$ represent the corresponding outputs. Although $ y_4$ is capable of capturing the temporal feature of the entire input sequence, $ y_1$  is restricted to capturing only the temporal information of $ x_1$.
	
	\item Spatial graphs of traffic data across different modes are inherently heterogeneous, ranging from fixed node graphs to grid graphs. The challenge lies in effectively integrating feature interactions across these disparate graphs. Furthermore, the varying scales of data across different modalities necessitate the performance of spatial-temporal alignment operations during the joint modeling process. As depicted in Figure \ref{Fig: BJFeature}, within the same mode of traffic, the changes in node inflow and outflow exhibit a similar pattern. However, as demonstrated in Figure \ref{Fig: Multimode_Feature}, the scale of flow changes varies significantly across different modes.
	
\end{itemize}

To address the above challenges, we introduce a novel multi-modal joint prediction framework, underpinned by a graph sparse attention mechanism and a bidirectional temporal convolutional network. Firstly, we implement data normalization across disparate scales, mapping them to a compatible data range. Subsequently, the graph sparse attention mechanism is employed to extract spatial features, while a shared-unique bidirectional temporal convolutional network module is utilized for the extraction of temporal features, which are then output via a multilayer perceptron prediction module. Finally, we perform separate normalization for data of varying patterns, reverting them to their original scale. The main contributions of this study are as follows:

\begin{itemize}
	
	\item For the extraction of spatial features, we employ the mechanism of graph and self-attention multiplication, which allows us to capture the time-varying features present in the adjacency graph structure, and thereby obtain spatial local features. To supplement this, we adopt the Top-U sparse self-attention mechanism, which enables the extraction of global spatial features across all nodes.
	
	\item For the extraction of temporal features, we incorporate a bidirectional temporal convolutional network. This network is capable of concurrently extracting temporal features from the input data in both forward and backward directions, thereby yielding a more comprehensive and enriched representation of temporal features. Building upon this, we construct shared-unique components of the BiTCN, which correspond respectively to the extraction of inter-modal and intra-modal temporal features.
	
	\item We propose a joint prediction framework that is easily scalable in both spatial and temporal dimensions. After extensive experiments on three real-world datasets, GSABT consistently achieves state-of-the-art results.
	
\end{itemize}

\section{Literature review}
\label{se:Literature review}

\subsection{Traffic prediction}

In statistical methods, ARIMA captures the autoregressive characteristics of time series data by linearly combining observational data, and introduces seasonal differences to model the seasonal changes in traffic flow \cite{van1996combining}. The Kalman filter estimates the state of traffic flow based on observational data and a system dynamics model \cite{okutani1984dynamic}. The Seasonal Autoregressive Integrated Moving Average model (SARIMA) accommodates seasonality and trends, while the Winters Exponential Smoothing model smooths historical data based on the weighted average method \cite{williams1998urban}. Although statistical models have a small number of parameters and strong interpretability, they require high-quality data, rely on assumptions of linear relationships, and may not be suitable for large-scale data.

In traditional machine learning methods, the adaptive multi-kernel support vector machine (AMSVM) uses a hybrid of Gaussian kernels and polynomial kernels, and combines this with the adaptive particle swarm algorithm to optimize the AMSVM parameters \cite{feng2018adaptive}. The improved KNN replaces the physical distance between road segments with an equivalent distance, uses a spatial-temporal state matrix to represent traffic status, and selects the nearest neighbor based on Gaussian weighted Euclidean distance \cite{cai2016spatiotemporal}.Gradient Boosting Decision Trees (GBDT) obtain the final result by iteratively training a series of decision trees and combining their prediction results with weights \cite{ma2017prioritizing}. While traditional machine learning methods are widely used, easy to implement, and highly computationally efficient, their ability to model complex spatial-temporal relationships is limited, and they can struggle with multi-step predictions.

Deep learning-based prediction methods have also been widely used in the field of spatial-temporal traffic data. DCRNN combines diffusion convolution with a gated recurrent unit to model the spatial-temporal features of traffic speed \cite{li2018diffusion}. STGCN employs a graph convolution network to model spatial features and a temporal gated convolution to model temporal features \cite{yu2017spatio}. ASTGCN integrates a spatial-temporal attention mechanism with spatial-temporal convolution, including graph convolution in the spatial dimension and convolution in the temporal dimension, to capture dynamic spatial-temporal features of traffic flow data \cite{guo2019attention}. Graph WaveNet proposes an adaptive adjacency matrix to capture hidden spatial features, combining graph convolution with a temporal convolution network \cite{wu2019graph}. GMAN uses spatial and temporal attention mechanisms along with a gated fusion mechanism to adaptively fuse spatial-temporal features \cite{zheng2020gman}. STSGCN employs a spatial-temporal synchronous graph convolutional module to model correlations, with multiple module layers designed to capture the heterogeneity of long-term spatial-temporal networks \cite{song2020spatial}. AGCRN \cite{bai2020adaptive} adopts a node adaptive parameter learning (NAPL) module to capture node patterns, and a data adaptive graph generation (DAGG) module to automatically determine dependencies between data. STTN \cite{xu2020spatial} uses a spatial transformer to model time-varying directed spatial correlations and a temporal transformer to facilitate long-term multi-step prediction. Bi-STAT \cite{chen2022bidirectional} implements a bidirectional spatial-temporal adaptive architecture for prediction, with one decoder performing present past recall tasks and the other decoder carrying out present future prediction tasks. USTAN \cite{long2022unified} constructs spatial and temporal neighborhood graphs to model spatial and temporal correlations, synchronously obtaining spatial-temporal correlations through self-attention mechanisms, and adaptively integrating external factors using gated fusion mechanisms. DTC-STGCN \cite{xu2023dynamic} proposes to capture traffic spatial features based on attention and a dynamic adjacency matrix, and utilizes LSTM to model temporal features. STMFFN \cite{wang2023spatial} designs a gated graph convolution module to capture spatial similarity and proposes a multi-scale attention module to obtain temporal dependence. MISTAGCN \cite{tao2023multiple} considers the correlation between recent, daily, and weekly periods, as well as external factors, and comprehensively adopts temporal attention mechanisms, spatial attention mechanisms, a graph convolutional network, and a temporal convolutional network for spatial-temporal feature extraction. TGAN \cite{wang2023trend} does not rely on a prior graph structure, but instead uses a trend space attention module to learn spatial features, and subsequently employs a pyramid attention module to learn both local and global temporal features. TIRE \cite{wang2023tyre} uses an attention mechanism to understand the correlation of all nodes, then employs a gating mechanism to control the fusion of near and far information, and finally uses a temporal convolutional network for prediction. However, these models tend to focus more on tasks within the same transportation mode, with less consideration given to the interaction of traffic feature information across different modes.

\subsection{Multimodal traffic prediction}

In multimodal traffic prediction, some models are based on the same spatial node division, but they consider different flow features, such as inflow and outflow. For instance, ST-ResNet \cite{zhang2017deep} takes into account external factors like weather and week information, and designs an end-to-end residual network structure to simulate the temporal closeness, period, and trend properties of crowd inflow and outflow. CCRNN \cite{ye2021coupled}  utilizes a coupling mechanism to associate the learnable adjacency graphs of different layers, and combines this with a GRU to generate prediction data. MVFN \cite{zhang2023multi} captures the complexity and spatial-temporal relationships of traffic demands by integrating information from multiple views. Given that the neural process is interpretable and provides probabilistic confidence for prediction results, TENP \cite{xu2023multi} is suitable for predicting pickups and returns in the bicycle-sharing system.

Some models conduct joint predictions for different traffic modes. For example, Cost-Net \cite{ye2019co} uses CNN and LSTM to jointly model the pickup and drop-off demand of taxis and shared bikes. MultiST \cite{wang2021learning} employs CNN and GRU to learn unique knowledge, while using a Recurrent Gaussian Cell to learn temporal dependence. KA2M2 \cite{li2021multi} uses knowledge adaptation with an attentive multi-task memory network to learn information about different modals, thereby improving demand prediction in sparse modals. MIX-MGC \cite{ke2021joint} employs regularization cross-task learning to model the connection of multi-graph convolutional networks (MGC), and uses multi-linear relations (MLR) to learn the weights of multiple networks to impose a prior tensor normal distribution. ST-MRGNN \cite{liang2022joint} uses a Multi-Relationship Graph Neural Network (MRGNN) to capture heterogeneous spatial dependencies across patterns, and integrates a temporal convolutional layer to jointly model heterogeneous spatial-temporal features. Res-Transformer \cite{yang2023short} can effectively capture the features of multiple traffic modals, such as taxis, buses, and subways, while the residual network ensures the stability of training. MBA-STNet \cite{miao2022mba} adopts a shared-private framework to learn multi-task features, introduces adversarial loss to reduce the redundancy of information in shared feature extraction, and uses Bayesian heterogeneous spatial-temporal modeling to alleviate data uncertainty.

\section{Preliminaries}

\subsection{Multimodal traffic system}

Assuming a multimodal traffic flow system consists of  $ M$ modes of transportation, each mode $ m$ consisting of $ N_{m}$ nodes as the basic unit. Assuming that each mode of transportation contains $ F$ features, such as inflow, outflow, etc, then the $ i$-th node at $ t$-th time period can be represented as an $ F$-dimensional vector $ x_{i}^{t} \in \mathbb{R}^{F}$. The traffic flow of the $ m$-th modal at all nodes in the $ t$-th time period can be represented as:

\begin{equation}
	X_{m}^{t} =\left \{x_{0}^{t}, x_{1}^{t}\dots ,x_{N_{m}}^{t}\right \}  
\end{equation}
where $ F$ is the feature of each node, and $ X_{m}^{t}\in \mathbb{R}^{N_{m}\times F}$.

All traffic modes are concatenated in the spatial dimension, and the traffic demand for the $ t$-th time period is:

\begin{equation}
	X_{M}^{t} = {\rm Concat} (X_{m}^{t}), \forall m 
\end{equation}
where $ N_{M}=\sum_{m=1}^{M} N_{m}$, $ N_{M}$ represents the total number of nodes for $ M$ modals, $ X_{M}^{t}\in \mathbb{R}^{N_{M}\times F}$.

\subsection{Multimodal traffic joint graph}

For the $ m$-th mode of transportation, we use $ G_m$ to represent the graph structure, $ G_m \in \mathbb{R}^{N_m \times N_m} $. If $ i$ and $ j$ nodes are connected, the value is assigned as 1, and otherwise, it is 0. The representation of nodes $ i$ and $ j$ in the $ m$-th graph is as follows:

\begin{equation}
	G_{m}(i,j) = \begin{cases}
		&1,   \text{ if $ i$ connects to $ j$} \\
		&0,   \text{ otherwise}
	\end{cases}
\end{equation}

Different modes of transportation generate heterogeneous traffic graphs. Therefore, we extend them along the diagonal and mask the remaining parts with 0-values, as shown in \ref{model}(b). This ensures local modeling of joint data in spatial extraction and is not affected by other modals. By extending the total $ M$ traffic graphs, a multi-modal graph can be obtained: 

\begin{equation}
	G_{M} = {\rm Extend} \left \{ G_{m}  \right \} , m\in M
\end{equation}

\subsection{Multimodal traffic joint prediction}

We define multimodal traffic joint prediction as: For a given set of multimodal traffic flow data from the past $ P$ time points, the task is to predict the data for the future $ Q$ time points.

\begin{equation}
	X_{M}^{t+1:t+Q} =\Gamma (X_{M}^{t-P+1:t}, G_M )
\end{equation}
where $ \Gamma$ is the mapping function that needs to be learned, $ G_M$ is a multi-modal traffic joint graph.$ X_{M}^{t+1:t+Q} \in \mathbb{R}^{Q \times N_{M} \times F}$, $ X_{M}^{t-P+1:t} \in \mathbb{R}^{P \times N_{M} \times F}$.

\section{Methodology }

\subsection{Model architecture}

\begin{figure*}[htbp]
	\centering
	\includegraphics[width=1.0\linewidth]{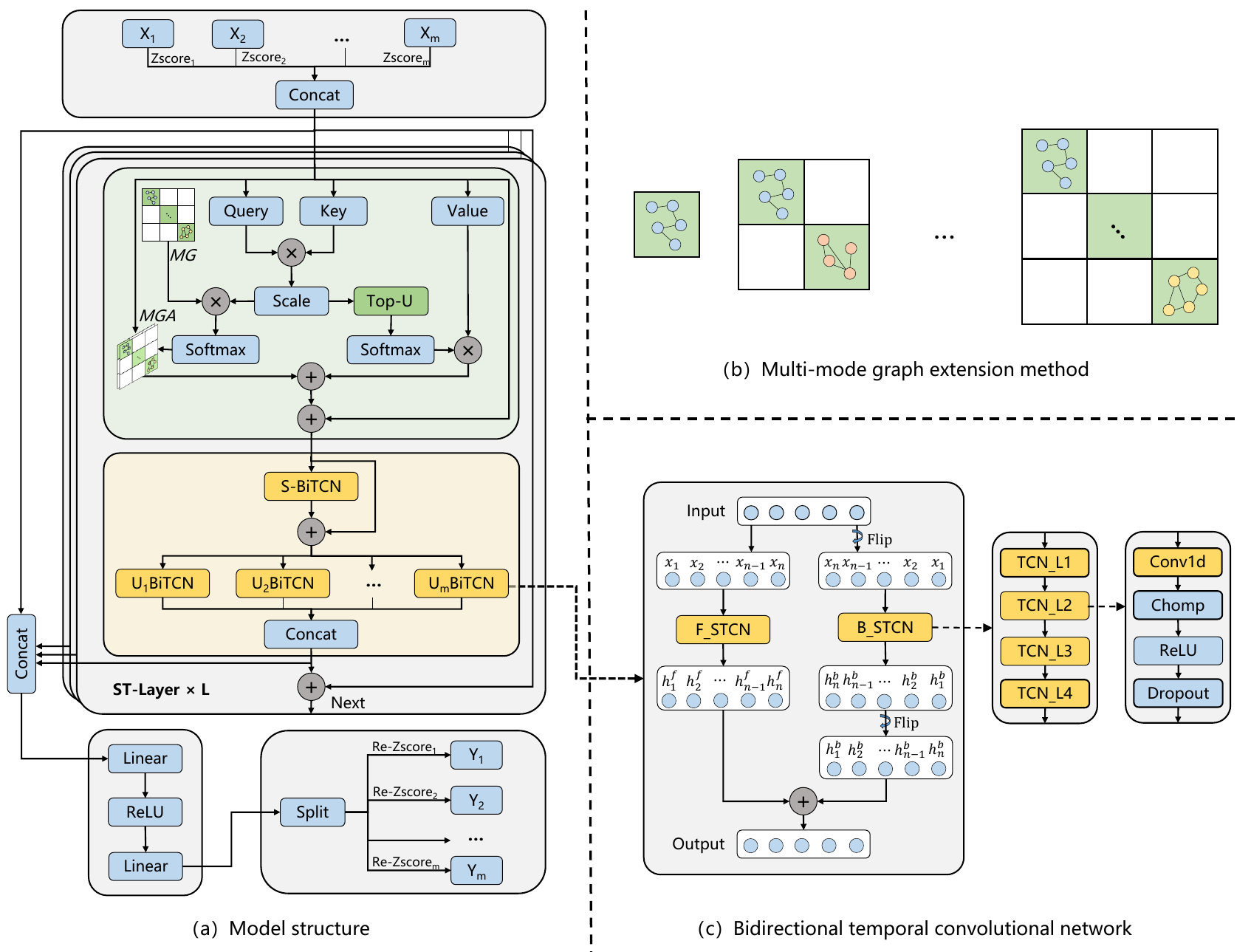}
	\caption{The architecture of GSABT.}
	\label{model}
\end{figure*}

Our proposed model, the Graph Sparse Attention with Bidirectional Temporal Convolutional Network (GSABT), is depicted in Figure \ref{model} (a). Firstly, the heterogeneous data types are normalized and fed into the spatial-temporal extraction layers. Then, residual connections are used to feed the data into a prediction layer of a Multilayer Perceptron (MLP) module. Finally, the segmented data is subjected to reverse normalization to restore it to its original scale.

We leverage a multimodal joint graph and self-attention weight multiplication to extract spatially related features. Subsequently, we employ the Top-U sparse attention mechanism to obtain inter-modal spatial features. In the temporal feature extraction phase, we utilize a Shared Bidirectional Temporal Convolutional Network (S-BiTCN) to capture the overall temporal features. Following this, we employ $ M$ Unique Bidirectional Temporal Convolutional Networks (U-BiTCN) to extract the specific temporal features associated with different modes of data.

The GSABT model demonstrates exceptional extensibility. In terms of spatial features, we expand the heterogeneous graph diagonally, filling the remaining areas with zeros to ensure the extraction of locally correlated information. For the temporal features, we use various U-BiTCN modules to extract temporal features across different modes of data.

\subsection{Graph Sparse Attention Mechanism}

Our spatial feature extraction technique is based on the self-attention mechanism, which is divided into a multimodal graph attention mechanism for local spatial feature extraction, and a sparse attention mechanism for global spatial feature extraction.

\subsubsection{Self-attention weight matrix}

To extract spatial features using CNN and GCN, the network layers need to be deepened to obtain distant features. Similar to a fully connected graph, the self-attention mechanism \cite{vaswani2017attention} enables the direct computation of pairwise relationships between different nodes by utilizing three functions: Query ($ Q$), Key ($ K$), and Value ($ V$) mappings:

\begin{equation}
	\begin{aligned}
		Q =XW_{Q} +b_{Q} \\
		K =XW_{K} +b_{K} \\
		V =XW_{V} +b_{V}
	\end{aligned}
\end{equation}
where $ W_{Q}$, $ W_{K}$ and $ W_{V}$ are weight matrices that are learnable, $ b_{Q}$, $ b_{K}$ and $ b_{V}$ are biases.

By performing operations on $ Q$ and $ K$, we obtain the dot product attention matrix:
\begin{equation}
	A_M = \frac{QK^{T} }{\sqrt{d_{k} } } 
	\label{Matrix}
\end{equation}
where $ d_{k}$ is the dimension of $ Q$, $ K$ and $ V$,  $ d_{k}  \in \mathbb{R} ^{P\times F}$.

\subsubsection{Multimodal graph attention mechanism}

GCN can extract features from non-Euclidean graphs. We expand the graph along the diagonal and mask the remaining parts with 0-values to ensure that the expanded graph focuses only on the spatial features within its own modality. We employ GCN and $ A_M$ to obtain the temporal features of adjacent nodes. By multiplying the prior graph with the self-attention matrix, and then passing through $ \rm Softmax$, a graph attention mechanism ($ G_{A}$) containing time-varying relationships can be obtained, and then we construct a two-layer GCN to extract spatial local features:

\begin{equation}
	G_{A}= {\rm Softmax}(G_{M}A_{M})
\end{equation}

\begin{equation}
	O_{G}(X)= {\rm ReLU}(G_{A}{\rm ReLU}(G_{A}XW_0) W_1)
\end{equation}
where $ W_0$ and $ W_1$ represent the learnable weights of the first and second layers, respectively.

\subsubsection{Sparse attention mechanism}

Considering the sparsity of traffic graphs, the use of a self-attention mechanism in the extraction of global spatial features will involve all nodes in the calculation \cite{liu2023sparsebev}. This can easily generate a long-tail effect and diminish the ability to extract meaningful features \cite{wang2021spatten}. To address this, we employ a sparse attention mechanism for global feature extraction, enabling cross-modal feature interaction. Specifically, we retain the Top-U values for each row of the attention matrix, fill the remaining entries with negative infinity, and apply $ \rm Softmax$ activation to obtain a sparse attention mechanism. By adding the outputs of the multi-modal graph attention and sparse attention, we can obtain the final spatial features.

\begin{equation}
	S_M(ij)=\begin{cases}
		& A_M(ij),    \text{ if } A_M(ij) \ge u_i \\
		& -\infty,   \text{ if } A_M(ij) <  u_i
	\end{cases}
\end{equation}
\begin{equation}
	O_{SA}(X) = {\rm Softmax}(S_M)V
\end{equation}
\begin{equation}
	O_{S} (X)= O_{G}(X) + O_{SA}(X)
\end{equation}
where $ u_i$ is the $ u$-th largest value of row $ i$.

\subsection{Bidirectional Temporal Convolutional Network}

\subsubsection{Temporal Convolutional Network}

The temporal convolutional layer in the model comprises four key components: a one-dimensional convolution with 'Chomp' operations to preserve causality; dilated convolution coefficients for interval sampling, which reduce computational complexity and broaden the receptive field; the ReLU function to counteract vanishing gradients; and a Dropout technique to prevent over-fitting by randomly omitting units during learning. Given a one-dimensional sequence $ X$ and convolutional kernel $ f$, the output of the expanded convolution at the $ i$-th time step $ H(i)$ can be expressed as:

\begin{equation}
	H(i)= \sum_{j=0}^{K-1} f(j)\cdot X_{i-d\cdot j} 
\end{equation}
where $ K$ is the size of the convolutional kernel, $ d$ is the dilated coefficient, $ f (j)$ is the $ j$-th element in the convolution kernel, and $ X_{i-d\cdot j} $ is the input sequence element corresponding to the convolution multiplication.

\subsubsection{Stacked Temporal Convolutional Network}

\begin{figure}
	\centering
	\includegraphics[width=0.8\linewidth]{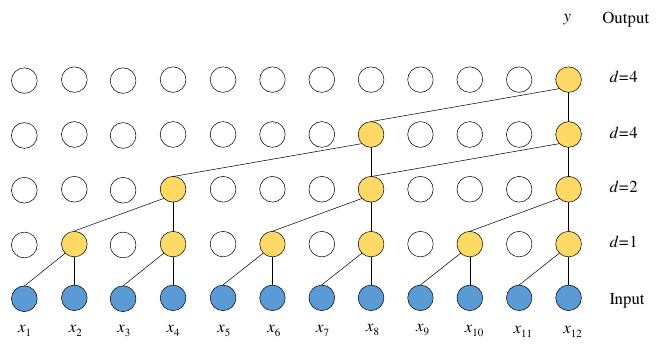}
	\caption{Stacked temporal convolutional network.}
	\label{fig: STCN}
\end{figure}

To increase the receptive field of the model, we have stacked four layers of a temporal convolutional network (STCN), as shown in \ref{fig: STCN}, with dilation coefficients of 1, 2, 4, and 4 respectively. By exponentially increasing the dilated coefficient, the model can significantly increase its receptive field without adding extra parameters or computational load. After four layers of dilated convolution, the receptive field of the model reaches 12, covering the fragment length $ P$ of the input data.

\subsubsection{Bidirectional Temporal Convolutional Network}

Each bidirectional temporal convolutional network (BiTCN) contains a forward STCN (F-STCN)  and a backward STCN (B-STCN). We assume that the entire F-STCN and B-STCN extraction modules are $  f_{FSTCN}$ and  $  f_{BSTCN}$ , respectively. In the extraction of B-STCN, the first step is to flip the input data $ X$ to obtain a reverse sequence, and then flip the output of $ f_{BSTCN}$ again to obtain reverse temporal features. We obtain the overall output of the bidirectional temporal features by adding forward and backward temporal features.

\begin{equation}
	H^{F}(X)= f_{FSTCN}(X)
\end{equation}
\begin{equation}
	H^{B}(X)= {\rm Flip}( f_{BSTCN}({\rm Flip} (X)))
\end{equation}
\begin{equation}
	O_{BT}(X) = H^{F}(X) + H^{B}(X)
\end{equation}
where $ f_{FSTCN}$ refers to the forward STCN extraction, $ f_{BSTCN}$ refers to the backward STCN                                                                                                                                                                                                                                                                                                                                                                                                                   extraction, $ O_{BT}(X) $ refers to bidirectional temporal features.

\subsubsection{Output of Temporal Features}

In temporal feature extraction, we stacked two layers of BiTCN, namely shared BiTCN (S-BiTCN) and unique BiTCN (U-BiTCN).

In S-BiTCN extraction, we use all node numbers $ N_{M}$ as convolutional channels, where all nodes can interact and achieve shared temporal feature extraction for all modalities.

In U-BiTCN extraction, we divide the data into different sub domains $ X_m$ based on the number of nodes in different modals, input them into corresponding $ U_m$-BiTCN, obtain independent temporal features of each modes, and then concatenate them along the spatial axis.

\begin{equation}
	O_{U_mT} = H^{F}(X_m) + H^{B}(X_m)
\end{equation}
\begin{equation}
	O_{T} = {\rm Concat} (O_{U_mT}), m \in M
\end{equation}

\subsection{MLP Module}
\label{se: prediction module}
The MLP module of our model is used to improve nonlinear prediction ability and is a fully connected neural network architecture \cite{vaswani2017attention}. The formula is as follows:

\begin{equation}
	\label{eq:MLP}
	{\rm MLP}(X)={\rm ReLU}(XW_{1}+b_{1})W_{2}+b_{2}
\end{equation}
where $ W_{1} $ and $ W_{2} $  are weight matrices, $ b_{1} $  and $ b_{2} $  are biases.

\subsection{Loss function}

We use mean absolute error (MAE) as the loss function, the formula is as follows:
\begin{equation}
	MAE = \frac{1}{n}\sum_{i=1}^{n} \left |y_{i}-y_{i}^{p}\right|  
	\label{eq:mae}
\end{equation}
where $ n$ is the total number of data, $ y_{i}$ is ground truth, $ y_{i}^{p}$ is the prediction data.

\section{Experiments}

\subsection{Datasets and Experiment Setup}

\subsubsection{Datasets}

The experiments conducted joint predictions on three real traffic datasets, collected from Beijing \cite{zhang2017deep} and New York \cite{ye2021coupled}, the detailed descriptions of the three datasets are shown in \ref{Dataset}.

\textbf{BJ Taxi:} To align the spatial-temporal data, we intercepted the traffic data of taxis in the Beijing area from January 8, 2016 to April 7, 2016. The area was divided into $ 16 \times 16$ according to a regular grid, with a total of 256 grids, including inflow and outflow traffic flow.

\textbf{NYC Bike:} The collection time was from April 1, 2016 to June 30, 2016. The collected information includes pick-up stations, drop-off stations, pick-up time, drop-off time, and trip time. Through the process of filtering, we retained the 256 most crucial nodes within the region.

\textbf{NYC Taxi:} The collection time was from April 1, 2016 to June 30, 2016. The collected information includes pick-up time, drop-off time, pick-up latitude and longitude, drop-off latitude and longitude, and trip distance. We formed 266 virtual nodes through clustering.

\begin{table}[]
	\centering
	\caption{Detailed description of three datasets}
	\begin{tabular}{cccc}
		\toprule
		Datasets       & BJ Taxi           & NYC Taxi           & NYC Bike           \\
		\midrule
		Time start    & 8/7/2016          & 1/4/2016           & 1/4/2016           \\
		Time end      & 7/4/2016          & 30/6/2016          & 30/6/2016           \\
		Time interval & 30min             & 30min              & 30min              \\
		Nodes         & 256               & 250                & 266                \\
		Feature       & Inflow/Outflow    & Pick up/ Drop off  & Pick up/ Drop off  \\
		\bottomrule
	\end{tabular}
	\label{Dataset}
\end{table}

\subsubsection{Baselines}

We used the following baseline methods to compare with the proposed GSABT.

\begin{itemize}
	
	\item \textbf{SVM:} SVM uses kernel functions to map low dimensional features to high-dimensional space, and minimizes the error between predicted and actual values by fitting the data and finding a function \cite{cortes1995support}.
	
	\item \textbf{KNN:} KNN treats historical data as points in a feature space, computes the distances between sample features during training, and identifies the k-nearest neighbor relationships. During prediction, the k samples with the closest features to the test set are selected as the nearest neighbors, and traffic prediction is performed using a weighted average approach \cite{cai2016spatiotemporal}.
	
	\item \textbf{BP:} The BP neural network enhances generalization by stacking multiple layers of neural networks and utilizing the ReLU activation function and Dropout technique.
	
	\item \textbf{LSTM:} LSTM adopts a gate mechanism to capture long-term dependencies in the sequence. At each time step, LSTM updates information based on the current input, hidden state of the previous time step, and cell state, which can capture important patterns and trends in the time series.
	
	\item \textbf{GWNET:} GWNET combines adaptive graph convolution with dilated causal convolution to capture spatial-temporal features \cite{wu2019graph}.
	
	\item \textbf{CCRNN:} CCRNN uses a coupling mechanism to associate the learnable spatial graph of different layers, and is integrated with the GRU to predict the spatial-temporal data \cite{ye2021coupled}.
	
	\item \textbf{MVFN:} MVFN employed a multi-perspective fusion approach for spatial-temporal prediction. It utilized GCN to extract neighboring spatial features, a linear attention mechanism to capture global spatial features, and a temporal convolutional network for temporal feature extraction \cite{zhang2023multi}.
	
\end{itemize}

\subsubsection{Implementation Details}

The time span of the three datasets we collected is 13 weeks. Specifically, we allocated the data from the initial 9 weeks as the training set, the data from the subsequent 2 weeks as the validation set, and the data from the final 2 weeks as the testing set. The deep learning model is implemented using Pytorch, while SVM and KNN are implemented using scikit-learn packages. The experimental platform is i7-8700, and the GPU is NVIDIA RTX3090.

The model parameters are set as follows. Batch size is set to 64, training epoch is set to 100, learning rate is 0.0005, feature 2 represents two different types of traffic, and Dropout is set to 0.1. We set the input historical data length $ P$ and the predicted data length $ Q$ to 12, the ST-Layer for basic comparison to 2, the value of Top-U to 16, and the number of layers for GCN to 2.

As a multimodal joint modeling, we constructed four types of spatial-temporal joint modeling and one type of single modal predict,  they are as follows:

\begin{itemize}
	\item \textbf{Single:} For each mode of traffic data, we make separate predictions, which is the traditional method for spatial-temporal traffic prediction.
	\item \textbf{NT-NB:} We conduct a joint prediction for different transportation modes under the same time and space, specifically for New York City Taxi and Bike traffic (NT-NB).
	\item \textbf{BT-NT:} We conduct a joint prediction for the same transportation mode under spatial-temporal heterogeneity, specifically for Beijing Taxi and New York City Taxi traffic (BT-NT).
	\item \textbf{BT-NB:} We conduct a joint prediction for different transportation modes under spatial-temporal heterogeneity, specifically for Beijing Taxi and New York City Bike traffic (BT-NB).
	\item \textbf{BT-NT-NB:} We conduct joint modeling for three types of data, specifically for Beijing Taxi, New York City Taxi, and New York City Bike traffic (BT-NT-NB).
	
\end{itemize}

\subsubsection{Evaluation Metrics}

We selected root mean squared error (RMSE), MAE \ref{eq:mae}, and Pearson correlation coefficient (PCC) as evaluation metrics.

\begin{align}
	&RMSE=\sqrt{\frac{1}{n}\sum_{i=1}^{n}(y_{i}-y_{i}^{p})^{2}} \\
	&PCC=\frac{ {\textstyle \sum_{i=1}^{n}(y_{i}^{p}-\bar{y^{p}})(y_{i}-\bar{y})}}{\sqrt{\sum_{i=1}^{n}(y_{i}^{p}-\bar{y^{p}})^{2}}\sqrt{\sum_{i=1}^{n}(y_{i}-\bar{y})^{2}} } 
	\label{eq:PCC}
\end{align}
where $ n$ is the total number of data, $ y_{i}$ is the ground truth, $ y_{i}^{p}$ is the predicted data, $ \bar{y}$ is the average of the ground truth, and $ \bar{y^{p}}$ is the average of the predicted data.

\subsection{Comparison with Baselines}

The comparison between our proposed model and the baseline models is presented in Table \ref{tab: Prediction}. Our model, denoted as GSABT, outperformed all other models across all tasks, as indicated by the bold figures. The best performing model among the comparison models is marked with an asterisk (*).

Although traditional machine learning algorithms such as SVM and KNN exhibited competitive performance, they require the independent construction of different models for each prediction step, node, and traffic feature. This leads to a total of $ Q \times N \times F$ models, which poses a significant challenge when applying these algorithms to large-scale spatial-temporal prediction tasks.

We further dissected the performance of our model from various perspectives, including dual data joint prediction, triple data joint prediction with augmentation coupling experiments, and single modal decoupling experiments.

\subsubsection{Dual data joint prediction analysis}

We conducted validation on the dual data experiment separately using NT-NB, BT-NT, and BT-NB configurations. The experimental results from the three groups were similar, so we selected the experimental results from the NT-NB group for presentation.

BP neural network can predict future time steps at once, but it is difficult to extract spatial features. LSTM considers nodes as features and can incorporate the interplay of all spatial-temporal data in predictions, resulting in good performance.

GWNET, and CCRNN can extract temporal and spatial features, thereby improving spatial-temporal prediction capabilities. MVFN can extract spatial-temporal features from multiple perspectives, preserving independent extraction of temporal features, resulting in better prediction performance and achieving the second best prediction effect in most tasks.

Compared to MFVN, GSABT's MAE and RMSE on NYC Taxi datasets decreased by 8.53\%, 8.55\%, and PCC increased by 0.63\%, respectively; On the NYC Bike datasets, MAE and RMSE decreased by 1.58\% and 0.29\% respectively, while PCC increased by 0.27\%.

\begin{table*}[]
	\centering
	\caption{Comparison of prediction results of various models under different tasks}
	
	\begin{tabular}{ccccccccccc}
		\toprule
		\multirow{2}{*}{\textbf{Task}} & \multirow{2}{*}{\textbf{Method}} & \multicolumn{3}{c}{\textbf{BJ Taxi}} & \multicolumn{3}{c}{\textbf{NYC Taxi}}  & \multicolumn{3}{c}{\textbf{NYC Bike}} \\
		& & \textbf{MAE}   & \textbf{RMSE} & \textbf{PCC} & \textbf{MAE} & \textbf{RMSE} & \textbf{PCC}    & \textbf{MAE} & \textbf{RMSE} & \textbf{PCC}  \\
		
		\midrule
		\multirow{2}{*}{\textbf{Independent}} 
		
		& SVM  & 19.9445 & 39.3861 & 0.9030 & 7.5877 & 14.1503 & 0.9216 & 2.2351 & 3.8685 & 0.6249   \\
		& KNN  & 19.3681 & 35.3938 & 0.9236 & 7.8034 & 14.1517 & 0.9238 & 2.2212 & 3.6408 & 0.6398 \\
		\midrule
		\multirow{8}{*}{\textbf{Single}}   
		& BP     & 22.7585 & 38.4255 & 0.8893 & 11.4315 & 21.6844 & 0.7797 & 2.4369 & 4.2611 & 0.4592 \\
		& LSTM   & 16.6747 & 30.4419 & 0.9396 & 6.2160 & 12.4432 & 0.9408 & 1.8091 & 3.0354 & 0.7697  \\
		& GWNET  & 22.0856 & 36.5160 & 0.9041 & 9.4977 & 19.3910 & 0.8478 & 1.7673 & 3.0611 & 0.7652  \\
		& CCRNN  & 17.5320 & 30.7842 & 0.9406 & 5.4979* & 9.5631*  & 0.9648* & 1.7404 & 2.8382 & 0.7934  \\
		& MVFN   & 15.6440* & 27.5710* & 0.9456* & 5.6665 & 9.7663  & 0.9642 & 1.6989* & 2.7981* & 0.8113*   \\
		& \textbf{GSABT}   & \textbf{14.5583} & \textbf{26.9832} & \textbf{0.9526} & \textbf{5.2843} 
		& \textbf{9.4357}  & \textbf{0.9663}  & \textbf{1.6773} & \textbf{2.7575}  & \textbf{0.8122} \\
		& IMP   & 6.94\%   & 2.13\%  & 0.74\% &3.89\% & 1.33\%  & 0.16\% & 1.27\% & 1.45\% & 0.11\%  \\
		\midrule
		\multirow{8}{*}{\textbf{NT-NB}} 
		& BP      &        &         &        & 11.3245 & 21.7170 & 0.7808 & 2.4717 & 4.3225 & 0.4386   \\
		& LSTM    &        &         &        & 5.4919  & 9.7616  & 0.9627 & 1.7097 & 2.8339 & 0.8026   \\
		& GWNET   &        &         &        & 7.1052  & 13.9363 & 0.9231 & 1.8886 & 3.3822 & 0.7113   \\
		& CCRNN   &        &         &        & 6.5882  & 14.7239 & 0.9136 & 1.7925 & 3.1662 & 0.7418   \\
		& MVFN    &        &         &        & 5.3976*  & 9.7162*  & 0.9633* & 1.6847* & 2.7249* & 0.8173*   \\
		& \textbf{GSABT}   & \textbf{} & \textbf{} & \textbf{}   & \textbf{4.9374}  & \textbf{8.8857} 
		& \textbf{0.9694} & \textbf{1.6582}   & \textbf{2.7171} & \textbf{0.8196}  \\
		& IMP    &   &     &        & 8.53\% & 8.55\%  & 0.63\%  & 1.58\%  & 0.29\% & 0.27\%          \\
		\midrule
		\multirow{8}{*}{\textbf{BT-NT}}        
		& BP    & 21.9536  & 37.7140 & 0.8900 & 11.5330 & 21.1477 & 0.7900 &        &        &        \\
		& LSTM  & 17.0353  & 31.7816 & 0.9348 & 5.8023  & 10.1977* & 0.9592* &        &        &        \\
		& GWNET & 16.5517  & 29.6698 & 0.9419 & 8.0317  & 16.1257 & 0.8959 &        &        &        \\
		& CCRNN & 14.5154*  & 26.9496 & 0.9486 & 6.6452  & 15.1037 & 0.9090 &        &        &         \\
		& MVFN  & 14.6548  & 26.4172* & 0.9527* & 5.7831*  & 10.2757 & 0.9581 &        &        &         \\
		& \textbf{GSABT}   & \textbf{12.9856} & \textbf{24.7254}  & \textbf{0.9591} & \textbf{5.1802} 
		& \textbf{9.3747}  & \textbf{0.9667}  & \textbf{} & \textbf{} & \textbf{}       \\
		& IMP   & 10.54\%  & 6.40\%  & 0.67\% & 10.43\% & 8.07\%  & 0.78\% &        &        &        \\
		\midrule
		\multirow{8}{*}{\textbf{BT-NB}}       
		& BP    & 21.5645  & 36.2280 & 0.9006 &  &  &  & 2.4822  & 4.3257   & 0.4328          \\
		& LSTM  & 17.9609  & 33.4011 & 0.9369 &  &  &  & 1.8439  & 3.1027   & 0.7601          \\
		& GWNET & 16.7849  & 30.6852 & 0.9349 &  &  &  & 2.0379  & 3.7221   & 0.6477          \\
		& CCRNN & 15.0654  & 28.4139 & 0.9437 &  &  &  & 1.8783  & 3.3448   & 0.7077          \\
		& MVFN  & 14.3996*  & 27.0108* & 0.9495* &  &  &  & 1.7838*  & 2.9048*   & 0.7926*          \\
		& \textbf{GSABT}   & \textbf{12.5708} & \textbf{24.5138} & \textbf{0.9580} & \textbf{}
		& \textbf{} & \textbf{} & \textbf{1.6919} & \textbf{2.7820} & \textbf{0.8117} \\
		& IMP   & 12.70\%  & 9.24\%  & 0.90\% &  &  &   & 5.15\%  & 4.23\%  & 2.41\%          \\
		\midrule
		\multirow{8}{*}{\textbf{BT-NT-NB}}   
		& BP    & 22.5323  & 38.0415 & 0.8878 & 11.6043 & 21.7616 & 0.7751 & 2.5005 & 4.3295 & 0.4160  \\
		& LSTM  & 17.1704  & 31.7610 & 0.9337 & 5.9499  & 10.7037 & 0.9555 & 1.7185* & 2.8124 & 0.8017  \\
		& GWNET & 17.2111  & 30.5729 & 0.9377 & 7.8109  & 16.0605 & 0.8960 & 2.0029 & 3.6654 & 0.6617  \\
		& CCRNN & 14.8886  & 27.3468 & 0.9486 & 7.1556  & 15.0942 & 0.9083 & 1.8480 & 3.2774 & 0.7245  \\
		& MVFN  & 13.9905*  & 25.9050* & 0.9541* & 5.6602*  & 9.9568*  & 0.9608* & 1.7198 & 2.7520* & 0.8094*  \\
		& \textbf{GSABT}   & \textbf{12.4635} & \textbf{23.7332}  & \textbf{0.9608} & \textbf{5.0059} & \textbf{8.8877} & \textbf{0.9690} & \textbf{1.6590} & \textbf{2.7058} & \textbf{0.8210} \\
		& IMP   & 10.91\%  & 8.38\%  & 0.70\% & 11.56\% & 10.74\% & 0.85\% & 3.46\% & 1.68\% & 1.44\%  \\
		\bottomrule
	\end{tabular}
	\label{tab: Prediction}
\end{table*}

\subsubsection{Augmented coupling experiment}

We conducted an augmented coupling study and conducted joint modeling of three datasets. In the joint modeling and prediction task of NT-NB-BT, on the BJ Taxi datasets, the MAE and RMSE of GSABT decreased by 10.91\% and 8.38\% respectively, while the PCC increased by 0.70\%; On the NYC Taxi datasets, the MAE and RMSE of GSABT decreased by 11.56\% and 10.74\% respectively, while the PCC increased by 0.85\%; On the NYC Bike datasets, the MAE and RMSE scores of GSABT decreased by 3.46\%, 1.68\%, while PCC increased by 1.44\%.

\subsubsection{Decoupling experiment}

In the decoupling experiment, we only model a single datasets each time without feature learning between modalities. GSABT can still achieve the best prediction performance. Compared with the best in the comparative model, on the BJ Taxi datasets, the MAE and RMSE of GSABT decreased by 6.94\% and 2.13\% respectively, while the PCC increased by 0.74\%; On the NYC Taxi datasets, the MAE and RMSE of GSABT decreased by 3.89\% and 1.33\% respectively, while the PCC increased by 0.16\%; On the NYC Bike datasets, the MAE and RMSE scores of GSABT decreased by 1.27\%, 1.45\%, while PCC increased by 0.11\%.

\subsection{Ablation experiment}

\subsubsection{Module ablation analysis}

In order to verify the role of model modules in feature extraction, we conducted ablation experimental analysis on the composition of each module of the GSABT model.

\begin{itemize}
	\item \textbf{W/O SA}: Eliminate sparse attention modules, so that the model can only extract local spatial features.
	\item \textbf{W/O A*GCN}: Cancel the multi-modal GCN module, so that the model can only extract global spatial features.
	\item \textbf{W/O A *}: Remove the GCN of Attention, so that the spatial local features of the model can only be influenced by fixed node weights and cannot recognize time-varying feature relationships.
	\item \textbf{W/O FSTCN}: Remove the forward stacked temporal convolutional network.
	\item \textbf{W/O BSTCN}: Remove the backward stacked temporal convolutional network.
\end{itemize}

\begin{figure*}
	\centering
	\includegraphics[width=0.9\linewidth]{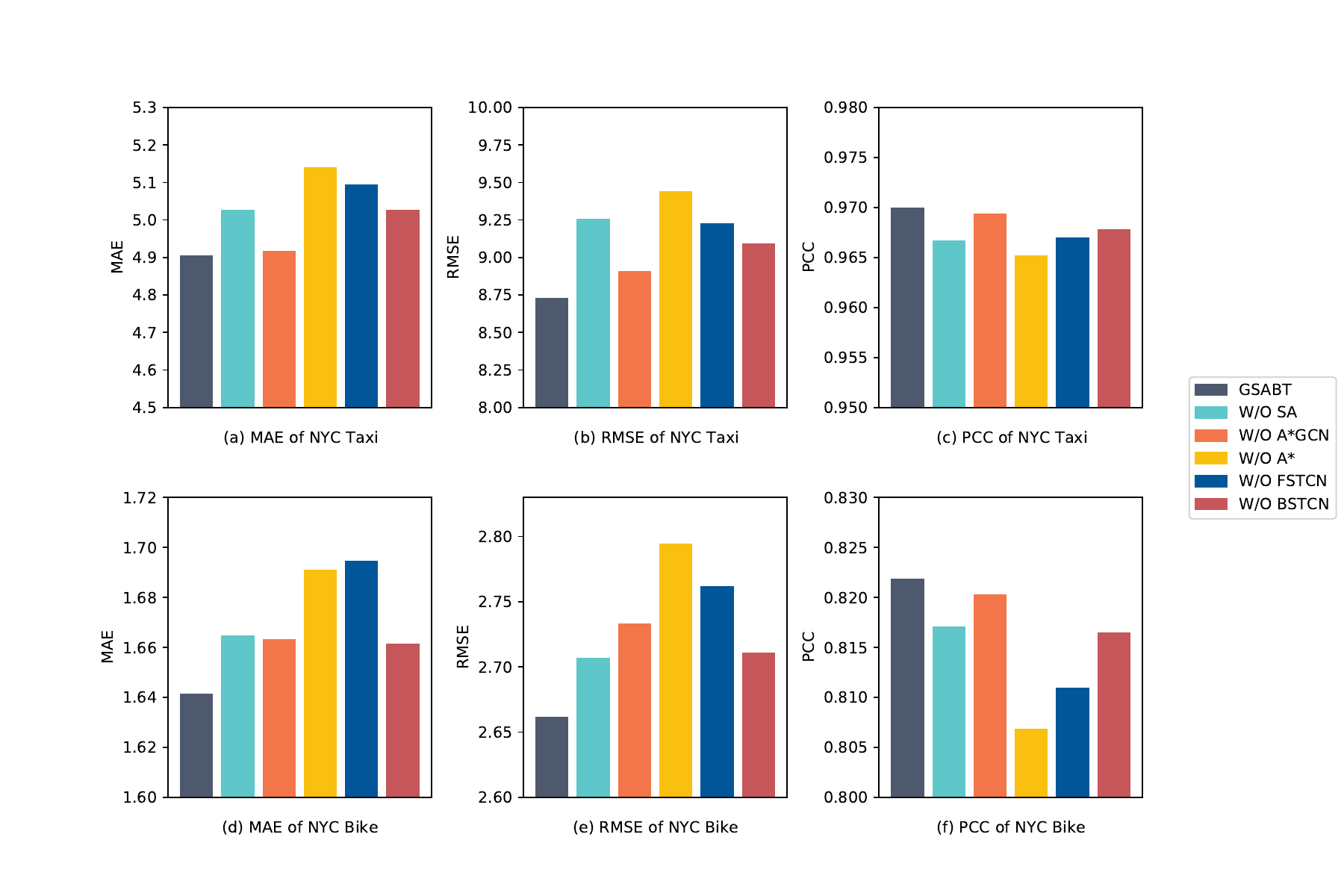}
	\caption{The ablation experiment of NT-NB joint prediction.}
	\label{fig: NT_NB_Abaltion}
\end{figure*}

We present the ablation experimental results of different modules predicted by NT-NB joint prediction as shown in \ref{tab: ablation} and \ref{fig: NT_NB_Abaltion}, and each module has played a promoting role in feature extraction. In the analysis of the spatial feature extraction module, we can see that the SA module can effectively assist the model in extracting global spatial features. When removing A * GCN, the model can only use sparse attention mechanism to extract spatial features, still achieving relatively high accuracy. In temporal feature extraction, the FSTCN and BSTCN modules are both helpful for temporal feature extraction. 

\begin{table}[]
	\centering
	\caption{Results of ablation experiments for each module}
	\begin{tabular}{ccccccc}
		\toprule
		\multirow{2}{*}{Method} & \multicolumn{3}{c}{NYC Taxi}  & \multicolumn{3}{c}{NYC Bike}  \\
		& MAE & RMSE & PCC & MAE & RMSE & PCC  \\
		\midrule
		\textbf{GSABT}  & \textbf{4.9057} & \textbf{8.7306} & \textbf{0.9700} & \textbf{1.6414} & \textbf{2.6619} & \textbf{0.8219}  \\
		W/O SA      & 5.0270 & 9.2559 & 0.9667 & 1.6647 & 2.7073 & 0.8171    \\
		W/O A*GCN   & 4.9175 & 8.9122 & 0.9694 & 1.6633 & 2.7332 & 0.8203    \\
		W/O A*      & 5.1409 & 9.4449 & 0.9652 & 1.6910 & 2.7947 & 0.8069    \\
		W/O FSTCN  & 5.0952 & 9.2306 & 0.9670 & 1.6946 & 2.7622 & 0.8110   \\
		W/O BSTCN  & 5.0279 & 9.0931 & 0.9678 & 1.6614 & 2.7109 & 0.8165    \\
		\bottomrule        	  
	\end{tabular}
	\label{tab: ablation}
\end{table}

\subsubsection{Analysis of ST-layers}

\begin{figure*}
	\centering
	\includegraphics[width=0.9\linewidth]{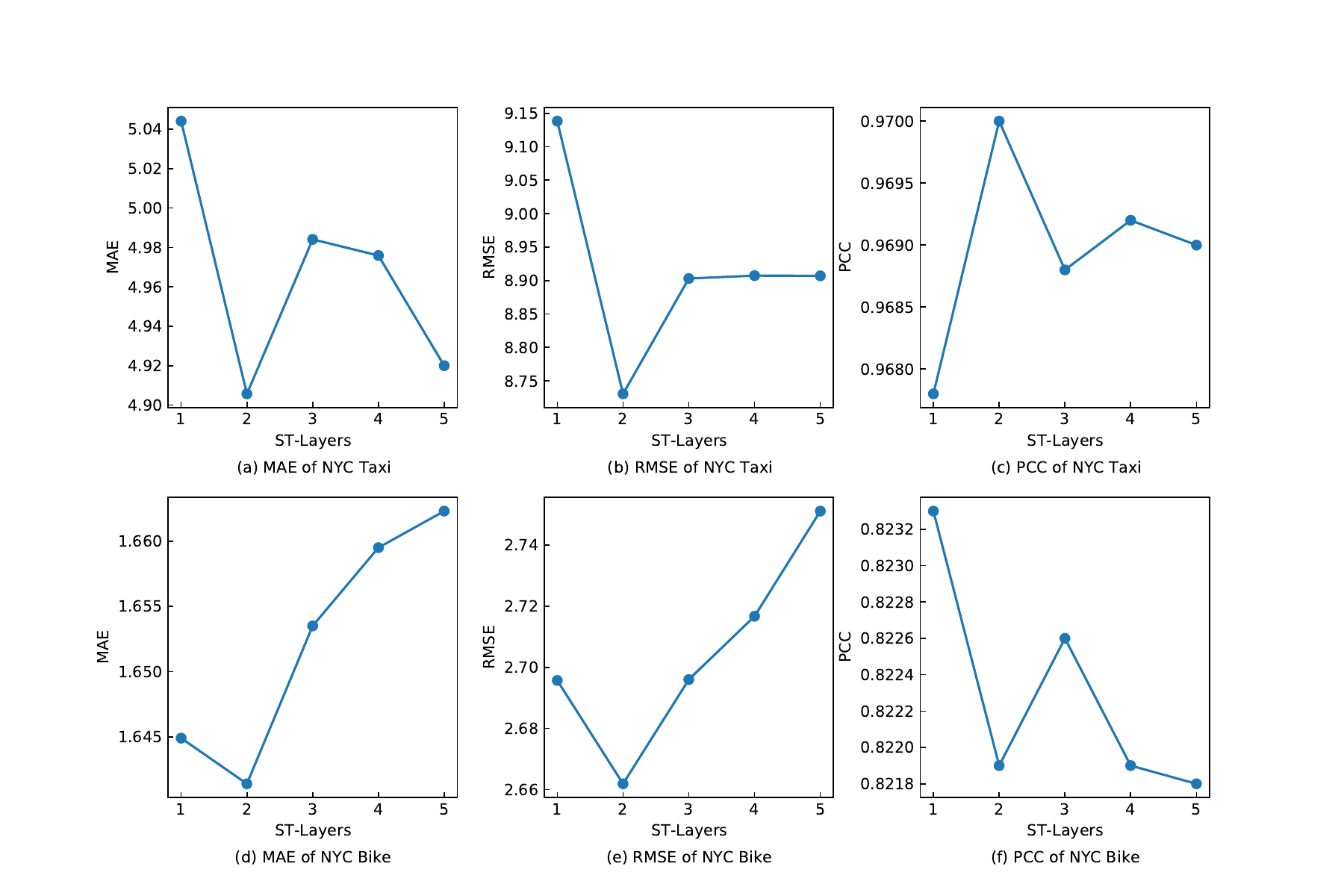}
	\caption{Error of different ST-layers.}
	\label{fig: STLayer_Ablation}
\end{figure*}

The experimental prediction results for different layers are shown in \ref{tab: st-layer} and \ref{fig: STLayer_Ablation}. We can see that when the number of layers is 2, a total of 5 indicators achieved the best prediction effect, and only the PCC indicators in the NYC Bike datasets achieved the best prediction effect at layer 1. Overall, when there are too few layers, the model cannot effectively extract spatial-temporal features, and excessive layers can also lead to a decrease in feature extraction ability.

\begin{table}[]
	\centering
	\caption{Prediction results for different layers}
	\begin{tabular}{ccccccc}
		\toprule
		\multirow{2}{*}{Layers} & \multicolumn{3}{c}{NYC Taxi} & \multicolumn{3}{c}{NYC Bike}   \\
		
		& MAE   & RMSE & PCC & MAE & RMSE & PCC    \\
		\midrule
		\textbf{1} & 5.0441  & 9.1382 & 0.9678 & 1.6449 & 2.6957 & \textbf{0.8233} \\
		2  & \textbf{4.9057} & \textbf{8.7306} & \textbf{0.9700} & \textbf{1.6414} & \textbf{2.6619} & 0.8219  \\
		3  & 4.9841 & 8.9030 & 0.9688 & 1.6535 & 2.6960  & 0.8226   \\
		4  & 4.9759 & 8.9073 & 0.9692 & 1.6595 & 2.7167  & 0.8219   \\
		5  & 4.9200 & 8.9070 & 0.9690 & 1.6623 & 2.7511  & 0.8218   \\
		\bottomrule    
	\end{tabular}
	\label{tab: st-layer}
\end{table}

\subsubsection{Top-U value analysis}

\begin{figure*}
	\centering
	\includegraphics[width=0.9\linewidth]{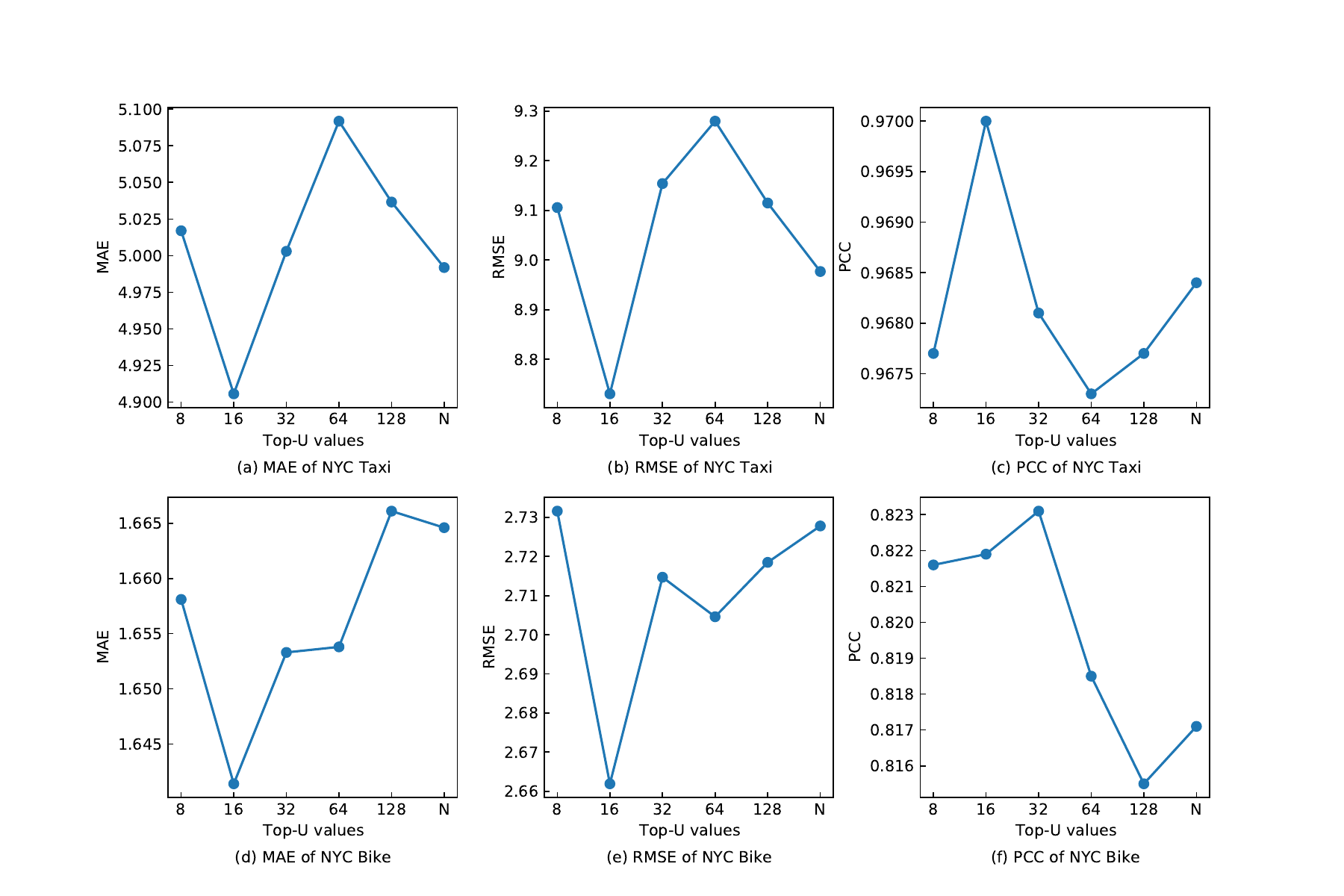}
	\caption{Model error under different values of Top-U.}
	\label{fig: Top-U Ablation}
\end{figure*}

The analysis of the impact of the sparse parameter U on various metrics is presented in \ref{tab: Top-U Ablation} and \ref{fig: Top-U Ablation}. When the value of U is set to 16, it leads in five key metrics, with the exception of the PCC metric for NYC Bike, where the optimal value is 32. In summary, if the value of U is too small, it is unable to encompass effective features, and when the value of U is too large, its ability to extract features diminishes.

\begin{table}[]
	\centering
	\caption{Prediction results for different Top-U values}
	\begin{tabular}{ccccccc}
		\toprule
		\multirow{2}{*}{Top-U} & \multicolumn{3}{c}{NYC Taxi}  & \multicolumn{3}{c}{NYC Bike}   \\
		& MAE & RMSE & PCC & MAE & RMSE & PCC \\
		\midrule
		8       & 5.0170 & 9.1058 & 0.9677 & 1.6581 & 2.7316 & 0.8216 \\
		16      & \textbf{4.9057} & \textbf{8.7306} & \textbf{0.9700} & \textbf{1.6414} & \textbf{2.6619} & 0.8219 \\
		32      & 5.0030 & 9.1541 & 0.9681 & 1.6533 & 2.7147 & \textbf{0.8231}   \\
		64      & 5.0918 & 9.2798 & 0.9673 & 1.6538 & 2.7046 & 0.8185   \\
		128     & 5.0366 & 9.1150 & 0.9677 & 1.6661 & 2.7185 & 0.8155   \\
		Full       & 4.9919 & 8.9770 & 0.9684 & 1.6646 & 2.7278 & 0.8171   \\    
		\bottomrule     
	\end{tabular}
	\label{tab: Top-U Ablation}
\end{table}

\subsubsection{Cross modal spatial feature extraction analysis}

\begin{figure}
	\centering
	\includegraphics[width=1.0\linewidth]{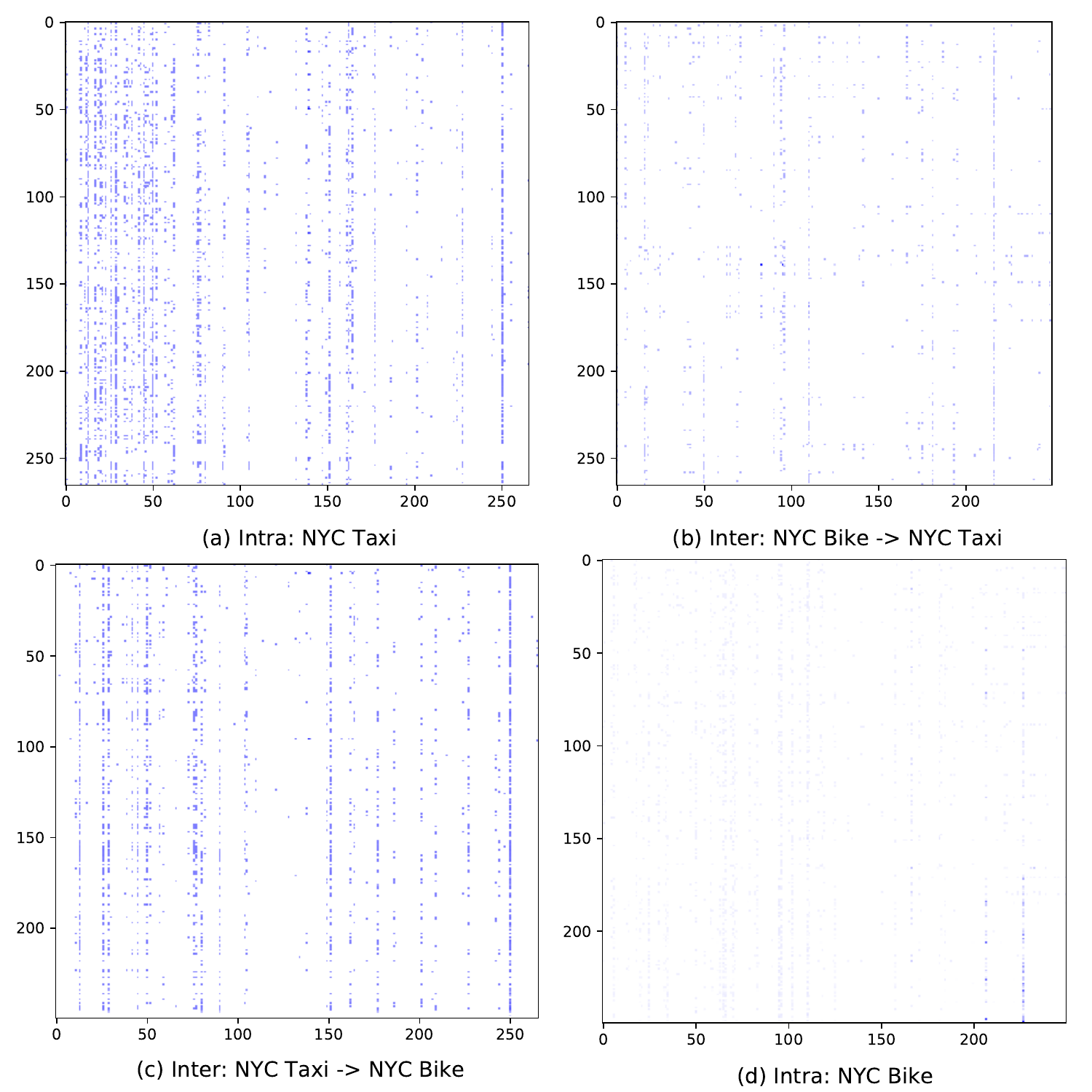}
	\caption{Cross modal feature extraction for NYC Taxi and NYC Bike}
	\label{fig:D_sparse}
\end{figure}

We studied the cross modal features of the model and visualized them. As shown in \ref{fig:D_sparse}, we demonstrate the sparse attention extraction mechanism of NT-NB joint modeling. We can see that both NYC Taxi and NYC Bike datasets can achieve complementary feature extraction. Among them, NYC Taxi has a higher attention weight, while NYC Bike shares less attention weight.

We conducted statistics on the weight distribution between different modalities, as shown in \ref{tab:NT-NB weight}. Among the nodes that contribute to NYC Taxi, NYC Taxi data contributed 3162 nodes, accounting for 74.30\%, while NYC Bike datasets contributed 1094 nodes, accounting for 25.70\%. Among the nodes that contribute to NYC Bike, the NYC Taxi datasets contributed 1791, accounting for 44.78\%, while the NYC Bike datasets contributed 2209 nodes, accounting for 55.23\%.  

\begin{table}[]
	\centering
	\caption{NT-NB weight distribution statistics}
	\begin{tabular}{ccccc}
		\toprule
		\multirow{2}{*}{Weight   statistics} & \multicolumn{2}{c}{NYC Taxi}   & \multicolumn{2}{c}{NYC Bike}     \\
		& Intra  Taxi & Bike-\textgreater Taxi & Taxi -\textgreater Bike & Intra Bike \\
		\midrule
		Weight nodes      & 3162     & 1094     & 1791     & 2209    \\
		Proportion        & 74.30\%  & 25.70\%  & 44.78\%  & 55.23\% \\
		\bottomrule     
	\end{tabular}
	\label{tab:NT-NB weight}
\end{table}

\section{Conclusion}

In this article, we propose a multimodal traffic spatial-temporal joint prediction network based on graph sparse attention mechanism and bidirectional temporal convolutional network(GSABT). Firstly, we use the graph sparse attention mechanism (GSA) to extract spatial features, multiply the weights of the spatial multimodal graph and the self-attention weights to obtain time-varying adjacency features, and obtain global spatial features entirely based on data-driven Top-U sparse attention mechanism. Secondly, we designed a bidirectional temporal convolutional network (BiTCN) to enrich temporal feature extraction in both forward and backward directions, based on BiTCN, we constructed a share-unique module for inter-modal and intra-modal temporal feature extraction. Finally, we have developed a highly scalable joint prediction framework that encompasses both spatial and temporal dimensions. Extensive experiments have been conducted on three real-world datasets, and GSABT consistently achieves state-of-the-art prediction results.

In the future, we plan to generalize the model to a wider range of multimodal data joint prediction and integrate it with technologies such as large language models to advance the development of traffic spatial-temporal prediction.

\section*{Declaration of competing interest}
The authors declare that they have no known competing financial interests or personal relationships that could have appeared to influence the work reported in this paper.

\section*{Acknowledgements}
This work was supported financially by the Science and Technology Planning Project of  Guangdong Province, grant number 2023B1212060029.

\balance
\bibliographystyle{elsarticle-num} 				
\bibliography{GSABT}

\end{document}